\documentclass[lettersize,journal]{IEEEtran}
\usepackage{amsmath,amsfonts}
\usepackage{algorithmic}
\usepackage{array}
\usepackage[caption=false,font=normalsize,labelfont=sf,textfont=sf]{subfig}
\usepackage{textcomp}
\usepackage{stfloats}
\usepackage{url}
\usepackage{verbatim}
\usepackage{graphicx}
\def\BibTeX{{\rm B\kern-.05em{\sc i\kern-.025em b}\kern-.08em
    T\kern-.1667em\lower.7ex\hbox{E}\kern-.125emX}}
\usepackage{balance}
\usepackage{xcolor}
\usepackage{cleveref}

\usepackage{multirow}
\usepackage{arydshln}
\usepackage{amssymb}
\usepackage{algorithmic}
\usepackage{algorithm}
\usepackage{comment} 

\begin{document}
\title{Set a Thief to Catch a Thief: Combating Label Noise through Noisy Meta Learning}
\author{Hanxuan Wang, Na Lu, Xueying Zhao, Yuxuan Yan, Kaipeng Ma, Kwoh Chee Keong, Gustavo Carneiro
\thanks{This work was supported by the National Natural Science Foundation of China under Grant U22B2036 and Grant 62476213. \textit{(Corresponding Author: Na Lu)}}

\thanks{Na Lu, Hanxuan Wang, Yuxuan Yan, and Kaipeng Ma are with the Department of Automation Science and Engineering, Xi’an Jiaotong University, China (email:lvna2009@mail.xjtu.edu.cn, {wanghanxuan1995, yan1611, 15958880}@stu.xjtu.edu.cn)}

\thanks{Xueying Zhao is with the School of Electrical Engineering, Xi’an Jiaotong University, China (email: zxy678@stu.xjtu.edu.cn)}

\thanks{Kwoh Chee Keong is with the College of Computing and Data Science, Nanyang Technological University, Singapore (e-mail: asckkwoh@ntu.edu.sg).}

\thanks{Gustavo Carneiro is with the Centre for Vision, Speech and Signal Processing, University of Surrey, UK (e-mail: g.carneiro@surrey.ac.uk).}
}

\markboth{Journal of \LaTeX\ Class Files,~Vol.~xx, No.~x, September~xxxx}%
{How to Use the IEEEtran \LaTeX \ Templates}

\maketitle

\begin{abstract}
Learning from noisy labels (LNL) aims to train high-performance deep models using noisy datasets. Meta learning based label correction methods have demonstrated remarkable performance in LNL by designing various meta label rectification tasks. However, extra clean validation set is a prerequisite for these methods to perform label correction, requiring extra labor and greatly limiting their practicality. To tackle this issue, we propose a novel noisy meta label correction framework STCT, which counterintuitively uses noisy data to correct label noise, borrowing the spirit in the saying ``\textbf{ \underline{S}}et a \textbf{ \underline{T}}hief to \textbf{ \underline{C}}atch a \textbf{ \underline{T}}hief''. The core idea of STCT is to leverage noisy data which is i.i.d. with the training data as a validation set to evaluate model performance and perform label correction in a meta learning framework, eliminating the need for extra clean data. By decoupling the complex bi-level optimization in meta learning into representation learning and label correction, STCT is solved through an alternating training strategy between noisy meta correction and semi-supervised representation learning. Extensive experiments on synthetic and real-world datasets demonstrate the outstanding performance of STCT, particularly in high noise rate scenarios. STCT achieves 96.9\% label correction and 95.2\% classification performance on CIFAR-10 with 80\% symmetric noise, significantly surpassing the current state-of-the-art.
\end{abstract}

\begin{IEEEkeywords}
Noisy labels, label correction, noisy meta learning, classification.
\end{IEEEkeywords}

\section{Introduction}
\IEEEPARstart{D}{eep} Deep learning has achieved great success in various fields, attributed to the availability of carefully annotated large scale datasets \cite{deep_learning1, deep_learning2, deep_learning3}. However, the collection of high quality datasets generally comes with high annotation cost and intensive human intervention, creating a significant obstacle to the development of deep learning.

Fortunately, the annotation cost issue can be mitigated through web crawling \cite{clothing1m} and crowdsourcing \cite{crowd_sourcing}. However, such low-cost datasets often contain a considerable amount of noisy labels, which may lead to severe overfitting of neural networks and performance degradation \cite{lnl_survey}. Therefore, it is crucial to investigate methods for training well-performing neural networks using datasets with noisy labels.

\begin{figure}[t]
  \centering
  \includegraphics[width=0.95\linewidth]{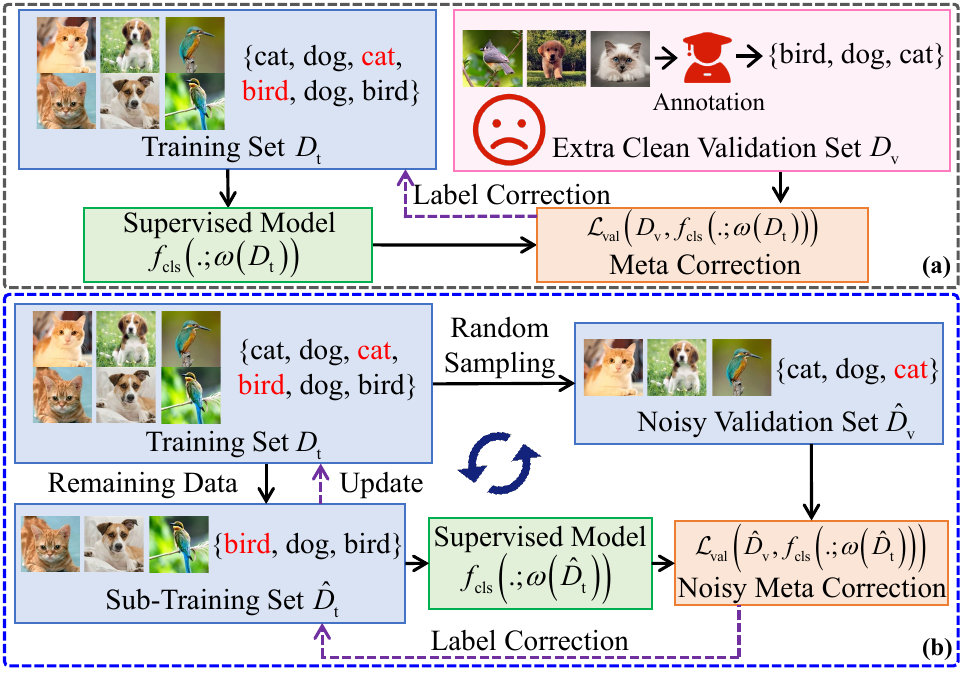}
   \caption{Comparison between (a) traditional meta learning based label correction framework and (b) STCT. STCT utilizes noisy validation sets for model performance evaluation, allowing label correction without using any extra clean data compared to traditional meta-learning methods.}
   \label{fig:comparison}
\end{figure}

\begin{figure}[t]
  \centering
  \includegraphics[width=0.83\linewidth]{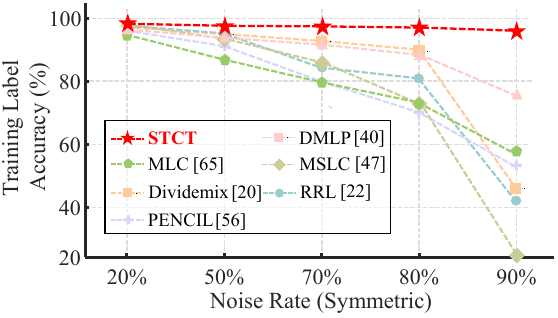}

   \caption{Comparison with other label correction methods on CIFAR-10 with symmetric noise. STCT's label correction capability significantly surpasses the competing methods.}
   \label{fig:label_correction_perform}
\end{figure}

To alleviate the adverse impact of noisy labels, many methods have been proposed these years, which can be roughly categorized into three classes: robust loss \cite{robust_loss1, robust_loss2, theory2, logitclip}, sample selection \cite{dividemix, c2d, feddiv}, and label correction \cite{dmlp, pencil, mlc}. Robust loss methods mitigate the neural networks' memorization of noisy labels by modifying loss functions, such as clipping logits \cite{logitclip}. Despite their theoretical robustness, these methods suffer from optimization difficulties, leading to underfitting on complex datasets \cite{underfitting}. Sample selection methods combat label noise by identifying clean samples based on training dynamics and employing semi-supervised methods to train the classifier. However, the selection criteria heavily rely on manual design, potentially limiting their generalization across various datasets \cite{josrc, lnl_survey}. Label correction methods, on the other hand,  aim to rectify noisy labels based on their geometric and probabilistic characteristics to improve the quality of training datasets. Among these, meta learning based label correction methods have demonstrated superior robustness against noisy labels \cite{dmlp, cto, mslc, mlc, mlnt}. In these methods, an extra clean validation set is introduced to evaluate model performance on the underlying clean distribution, and a series of meta tasks are formulated based on these data. The meta tasks treat noisy labels as hyper-parameters and correct them through a nested bi-level optimization, as shown in \cref{fig:comparison}(a). Despite achieving good performance, the need for a clean validation set inevitably introduces additional annotation cost, which contradicts the settings of the LNL problem and greatly limits the practicality of meta learning based label correction methods. Although some methods \cite{noisy_val1, noisy_val2} tried to generate a validation set using pseudo-labels or sample selection techniques from the training data to replace the extra clean data, these manually designed rules inevitably result in biased estimation of the clean distribution, leading to limited performance \cite{theory1}.

To address this issue, we propose ``\textbf{ \underline{S}}et a \textbf{ \underline{T}}hief to \textbf{ \underline{C}}atch a \textbf{ \underline{T}}hief'' (STCT) method, which is a novel noisy meta label correction framework that counterintuitively uses noisy data to correct label noise. We prove the optimal classifiers on a noisy dataset and its clean counterpart are consistent and thus noisy validation set can be directly used for label correction. Therefore, a noisy validation set, sharing the same distribution as the training data, can evaluate the model performance on the clean distribution, thereby enabling label correction through meta learning. In this way, STCT can eliminate the reliance on extra clean data, improving its practicality. To achieve better label correction, the bi-level optimization objective in STCT is solved through an alternating training strategy between noisy meta correction (NMC) and semi-supervised representation learning (SRL). Leveraging the encoder's representation learning capability, NMC constructs a linear model in the embedding space and utilizes the noisy validation set to evaluate model performance, thereby enabling label correction. SRL selects clean samples from the corrected training labels, enhancing the encoder's performance through semi-supervised learning. These two steps mutually reinforce each other, forming a positive feedback loop. Experimental results on several datasets demonstrate that STCT exhibits superior performance on label correction (see \cref{fig:label_correction_perform}) and classification. Our contributions can be summarized as follows:

\begin{itemize}
\item{We propose a novel noisy meta label correction framework, STCT, which uses noisy data to correct label noise, eliminating the reliance on extra clean validation set.}
\item{We further solve the bi-level optimization in STCT through an alternating training strategy between noisy meta correction and semi-supervised representation learning, forming a positive feedback loop to improve model performance.}
\item{Extensive experiments on both synthetic and real-world datasets demonstrate that STCT achieves superior label correction and classification performance, especially under high noise rate conditions.}
\end{itemize}

\section{Related Works}
\noindent This section introduces LNL methods related to STCT.

\subsection{Sample Selection Methods}

This category of methods mitigates the adverse effects of label noise by identifying noisy samples from the contaminated training datasets. Various dynamic learning features, such as training loss, can serve as criteria for sample selection \cite{dividemix, mild}. After distinguishing between clean and noisy data, a robust classifier can be trained by assigning smaller weights to noisy samples \cite{reweight1,reweight2, reweight3, reweight4, reweight5} or by discarding noisy labels and converting the problem into a semi-supervised learning task \cite{dividemix, c2d, mild, codis, feddiv}. Despite the remarkable performance, these methods heavily rely on manually designed sample selection rules, which may limit their generalization across different datasets \cite{lnl_survey}.

\subsection{Label Correction Methods}

These methods refurbish training labels to prevent neural networks from overfitting to noisy data. To achieve this, some methods use neural network predictions as pseudo labels to update noisy labels in the training data \cite{pencil, dividemix, label_correction1, label_correction2}. Another line of works employ unsupervised contrastive learning to obtain robust representations and perform label correction based on the geometric properties of the training data \cite{tcl, rrl, label_correction3}. For instance, Li et al. applied a smoothness constraint on neighboring samples based on geometric relationships in the embedding space to correct noisy labels \cite{rrl}. Unfortunately, these geometric or probabilistic property based label correction methods generally exhibit severe performance degradation in high noise rate scenarios \cite{label_correction3, label_correction4}.

\subsection{Meta Learning Based Methods}
With the core idea of learning to learn \cite{meta_learning_survey}, meta learning based LNL methods take sample weights \cite{meta_weight1, meta_weight2, meta_weight3} or training labels \cite{mlc, mlnt, mslc, dmlp, cto} as hyper-parameters, and learn appropriate hyper-parameters through designing different meta tasks to alleviate the impact of noisy labels. Among these, meta learning based label correction methods have demonstrated impressive results. For example, DMLP proposed a non-nested framework which trained a robust classifier through decoupling the label correction process into label-free representation learning and a linear meta label purifier \cite{dmlp}. Although effective, these methods require a clean validation set to assess model performance on the clean distribution, which incurs extra annotation cost and contradicts the settings of the LNL problem. What's more, some studies attempted to generate a validation set using pseudo labels or sample selection techniques. For example, FSR \cite{noisy_val1} proposed using the historical gradient information to select high-confidence clean samples, while FaMUS \cite{noisy_val2} constructed a clean validation set based on training loss. Despite the fact that a pseudo-clean validation set can be collected through these methods, the manually designed rules may introduce a biased estimation of the clean distribution, limiting model performance \cite{theory1}.

Unlike these existing meta learning based methods, our STCT proposes a novel noisy meta label correction framework which uses noisy data as the validation set to combat label noise, eliminating the reliance on extra clean data without introducing biased estimation.

\begin{figure*}[]
    \centering
    \includegraphics[width=1.0\textwidth]{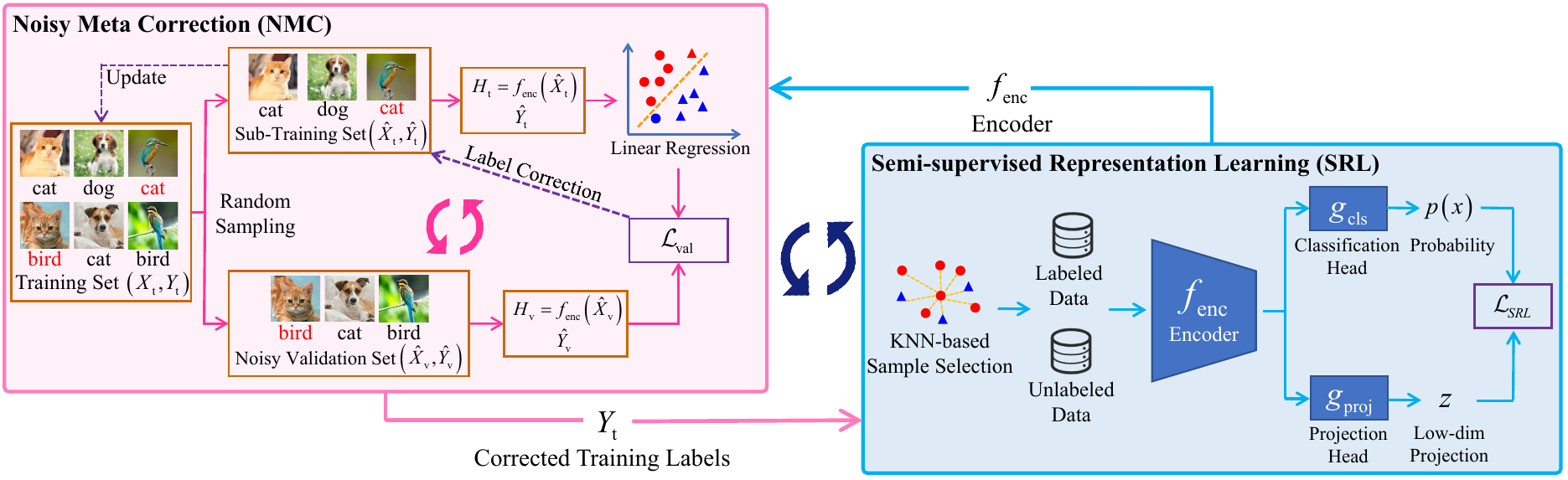}
    \caption{The overall framework of STCT. NMC constructs a linear model in the embedding space and uses the sampled noisy validation set to evaluate model performance, enabling label correction. SRL selects clean samples from the corrected training labels and uses semi-supervised learning to improve the inter-class separability of the embedding features extracted by the encoder.}
    \label{fig:diagram_method}
\end{figure*}
 
\section{Noisy Meta Label Correction Framework}

\noindent We use uppercase letters to represent matrices, bold lowercase letters to represent vectors, and scalars are in lowercase letters. Let $D_{\mathrm{t}}=\left\{ \left(\boldsymbol{x_{i}},y_{i} \right) \right\}^{n}_{i=1}$ denote a noisy label contaminated dataset with $n$ samples, where $\boldsymbol{x_i}$ and $y_i \in \{1, 2, \dots, C \}$ represent the feature vector and label of the $i$-th sample, and $C$ is the number of classes.

Taking the noisy labels as hyper-parameters, a typical meta learning based label correction framework can be formulated as a bi-level optimization problem between the weights of the classification model and the hyper-parameters. The noisy labels can be corrected through solving
\begin{align}
\small
\begin{matrix}
\min\limits_{{y_\mathrm{t}}} & \mathbb{E}_{\left( \boldsymbol{x_{\mathrm{v}}},y_{\mathrm{v}} \right)\in D_{\mathrm{v}}} \mathcal{L}_{\mathrm{val}}\left(f_{\mathrm{C}}\left(\boldsymbol{x_{\mathrm{v}}}; \boldsymbol{\omega^{\ast}} \right), y_{\mathrm{v}} \right),  \\
\textbf{s.t.} &  \boldsymbol{\omega^{\ast}} = \mathop {\arg \min }\limits_{\boldsymbol{\omega}} \mathbb{E}_{\left( \boldsymbol{x_{\mathrm{t}}},y_{\mathrm{t}} \right)\in D_{\mathrm{t}}} \mathcal{L}_{\mathrm{train}}\left(f_{\mathrm{C}}\left(\boldsymbol{x_{\mathrm{t}}}; \boldsymbol{\omega} \right), y_{\mathrm{t}} \right),  \\
\end{matrix}
\end{align}
where $D_{\mathrm{v}}$ is the validation set, $f_{\mathrm{C}}(.;\boldsymbol{\omega})$ is a classification model parametrized with $\boldsymbol{\omega}$, $\mathcal{L}_{\mathrm{train}}$ and $\mathcal{L}_{\mathrm{val}}$ are the loss functions on $D_{\mathrm{t}}$ and $D_{\mathrm{v}}$. To achieve label correction, $\mathcal{L}_{\mathrm{val}}$ should have the capability to evaluate the performance of $f_{\mathrm{C}}$ on the underlying clean distribution corresponding to $D_{\mathrm{t}}$.  As a result, previous studies introduced a clean validation set to approximate the clean distribution \cite{mlc, mlnt, mslc, dmlp, cto, meta_weight1, clean_val1}. However, the utilization of this clean validation set incurs additional annotation cost, greatly limiting the practicality of these methods.

To address this, a novel meta label correction framework is proposed which uses noisy data randomly sampled from the training set as the validation set, thereby eliminating the reliance on additional annotations. The use of a noisy validation set to correct noisy labels is like the saying ``set a thief to catch a thief"; hence, this framework is named as STCT. The core idea is that if a noisy validation set shares the same distribution as the training data, it can serve to evaluate the model's performance on the corresponding clean distribution. Inspired by \cite{theory1, theory2, theory3}, this idea is supported by the following two theorems:

\noindent\textbf{Theorem 1} Assume that the training set $D_{\textrm{t}}$ and its corresponding clean dataset $D_{\textrm{c}}$ are sampled from a noisy distribution $\tilde D_{\textrm{t}}$ and a clean distribution $\tilde D_{\textrm{c}}$, respectively. $T$ is the noise transition matrix, where $ T_{i,i}>\text{max}_{j\in \{1,2,...,C\},j\neq i}T_{i,j}$ for all $i\in \{1,2,...,C\}$, ensuring that the correct pattern can be distinguished from the noisy labels. Then, the optimal classifiers on $\tilde D_{\textrm{t}}$ and $\tilde D_{\textrm{c}}$ are consistent, satisfying
\begin{equation}
\small
\begin{array}{l}
{\rm{Ac}}{{\rm{c}}_{{{\tilde D}_\mathrm{t}}}}\left( {f_{{{\tilde D}_\mathrm{t}}}^*} \right) = {\rm{Ac}}{{\rm{c}}_{{{\tilde D}_\mathrm{t}}}}\left( {f_{{{\tilde D}_\mathrm{c}}}^*} \right) = \sum\limits_{i = 1}^C {\Pr \left[ {Y = i} \right]{T_{i,i}}}, \\
{\rm{Ac}}{{\rm{c}}_{{{\tilde D}_\mathrm{c}}}}\left( {f_{{{\tilde D}_\mathrm{t}}}^*} \right) = {\rm{Ac}}{{\rm{c}}_{{{\tilde D}_\mathrm{c}}}}\left( {f_{{{\tilde D}_\mathrm{c}}}^*} \right) = 1,
\end{array}
\end{equation}
where $f_D^*$ represents the optimal classifier on distribution $D$, and ${\rm{Ac}}{{\rm{c}}_D}\left( f \right)$ denotes the accuracy of classifier $f$ on distribution $D$. The proof of Theorem 1 is provided in Appendix A. Theorem 1 establishes the connection between the noisy and clean distributions, demonstrating that the noisy distribution can serve as a validation set to evaluate the model's performance on the clean distribution. However, since the noisy distribution is inaccessible, it is essential to analyze how effectively a noisy validation set with limited samples can evaluate model performance.

\noindent\textbf{Theorem 2} Considering a noisy validation set $D_{\mathrm{v}}$ with $n_{\mathrm{v}}$ samples is sampled from to the noisy distribution $\tilde D_{\mathrm{t}}$. Using Hoeffding's inequality, for any $\epsilon>0$, there exists
\begin{equation}
\small
\Pr \left[ {\left| {{{\hat {\cal R}}_{{D_\mathrm{v}}}}\left( f \right) - {\cal R_{{\tilde D_\mathrm{t}}}}\left( f \right)} \right| \ge \epsilon } \right] \le 2\exp \left( { - 2{n_\mathrm{v}}{\epsilon ^2}} \right),
\end{equation}
where ${\hat R_{{D_\mathrm{v}}}}\left( f \right) = \frac{1}{{{n_\mathrm{v}}}}\sum\limits_{i = 1}^{{n_\mathrm{v}}} {\ell \left( {f\left( \boldsymbol{{{x_i}}} \right),{y_i}} \right)}$ is the empirical error, ${\cal R}\left( f \right) = {\mathbb{E}_{\left( {\boldsymbol{x},y} \right) \sim {{\tilde D}_t}}}\left[ {\ell \left( {f\left( \boldsymbol{x} \right),y} \right)} \right]$ denote the generalization error on the noisy distribution, and $\ell(.,.)$ is 0-1 loss. 

Theorem 2 indicates that the larger the sample size of the noisy validation set $D_\mathrm{v}$, the more accurate the evaluation of the model's performance on the noisy distribution $\tilde D_{\mathrm{t}}$.

Combining Theorem 1 and Theorem 2, we can conclude that, with enough data, a noisy validation set, which follows the noisy distribution, can evaluate model performance on the corresponding clean distribution. As a result, using a noisy validation set as a substitute for the extra clean validation set is reasonable in traditional meta learning based label correction frameworks. However, the noisy distribution is invisible to the training process. Alternatively, a noisy validation set can be randomly sampled from the training set $ D_{\mathrm{t}}$, dividing the training data into a sub-training set $\hat D_{\mathrm{t}}$ and a noisy validation set $\hat D_{\mathrm{v}}$. $r=n_{\mathrm{v}}/n$ is the sampling rate. Then, the bi-level optimization of STCT is formulated as
\begin{align}
\small
\begin{matrix}
\min\limits_{{y_\mathrm{t}}} & \mathbb{E}_{\left( \boldsymbol{x_{\mathrm{v}}},y_{\mathrm{v}} \right)\in \hat D_{\mathrm{v}}} \mathcal{L}_{\mathrm{meta}}\left(f_{\mathrm{C}}\left(\boldsymbol{x_{\mathrm{v}}}; \boldsymbol{\omega^{\ast}} \right), y_{\mathrm{v}} \right),  \\
\textbf{s.t.} &  \boldsymbol{\omega^{\ast}} = \mathop {\arg \min }\limits_{\boldsymbol{\omega}} \mathbb{E}_{\left( \boldsymbol{x_{\mathrm{t}}},y_{\mathrm{t}} \right)\in \hat D_{\mathrm{t}}} \mathcal{L}_{\mathrm{train}}\left(f_{\mathrm{C}}\left(\boldsymbol{x_{\mathrm{t}}}; \boldsymbol{\omega} \right), y_{\mathrm{t}} \right).  \\
\end{matrix}
\label{eq:stct_optimization}
\end{align}
In fact, only solving \cref{eq:stct_optimization} would leave the labels in $\hat D_{\mathrm{v}}$ uncorrected. To address this issue, we update the corrected labels in the training set $D_{\mathrm{t}}$ and resample a new sub-training set and a new noisy validation set to solve \cref{eq:stct_optimization} once again. Theoretically, by repeating the random sampling for $\frac{\text{ln}\left( 1-\text{exp}\left( \text{ln}\left( \beta \right)/n \right) \right)}{\text{ln}\left( 1-r \right)}$ times (see the proof in Appendix B), at least $\beta$ percent of the training labels in $D_{\mathrm{t}}$ can be covered.

\section{Solving STCT}
In practice, directly solving \cref{eq:stct_optimization} is highly challenging, as it requires frequent searches for the optimal classifier based on the current sub-training set during the correction of noisy labels, resulting in unacceptably large computational cost. Inspired by \cite{dmlp}, the bi-level optimization of STCT can be simplified with a non-nested solving strategy, which employs a pre-trained encoder to extract representative embedding features, enabling a simple risk estimation function (e.g., linear model) to serve as the classifier in constructing $\mathcal{L}_{\mathrm{train}}$, thus reducing the computational cost. Therefore, the bi-level optimization can be fixed into
\begin{align}
\small
\begin{matrix}
\min\limits_{{y_\mathrm{t}}} & \mathbb{E}_{\left( \boldsymbol{x_{\mathrm{v}}},y_{\mathrm{v}} \right)\in \hat D_{\mathrm{v}}} \mathcal{L}_{\mathrm{meta}}\left(f_{\mathrm{C}}\left(\boldsymbol{h_{\mathrm{v}}}; \boldsymbol{\omega^{\ast}} \right), y_{\mathrm{v}} \right),  \\
\textbf{s.t.} &  \boldsymbol{\omega^{\ast}} = \mathop {\arg \min }\limits_{\boldsymbol{\omega}} \mathbb{E}_{\left( \boldsymbol{x_{\mathrm{t}}},y_{\mathrm{t}} \right)\in \hat D_{\mathrm{t}}} \mathcal{L}_{\mathrm{train}}\left(f_{\mathrm{C}}\left(\boldsymbol{h_{\mathrm{t}}}; \omega \right), y_{\mathrm{t}} \right),  \\
& \boldsymbol{h_{\mathrm{t}}} = f_{\mathrm{enc}}\left( \boldsymbol{x_{\mathrm{t}}}; \boldsymbol{\theta^{\ast}} \right), \\
& \boldsymbol{h_{\mathrm{v}}} = f_{\mathrm{enc}}\left( \boldsymbol{x_{\mathrm{v}}}; \boldsymbol{\theta^{\ast}} \right), \\
& \boldsymbol{\theta^{\ast}} \gets F_{\mathrm{Pretrain}}.
\end{matrix}
\label{eq:stct_optimization_decompose}
\end{align}
where $f_{\mathrm{enc}}$ is the encoder and $F_{\mathrm{Pretrain}}$ represents the pre-training method. We implicitly assumes that each class is linearly separable within the embedding space, as only under this condition can a linear model replace neural networks. The quality of the embedding features is highly related to the pre-training of the encoder. Therefore, solving \cref{eq:stct_optimization_decompose} can be decoupled into  representation learning ($F_{\mathrm{Pretrain}}$) and label correction. However, due to the presence of noisy labels, it is impractical to use supervised representation learning as $F_{\mathrm{Pretrain}}$. Therefore, in previous studies \cite{dmlp, tcl, label_correction3}, the encoder is primarily pre-trained through unsupervised contrastive learning \cite{simclr} or by directly adopting pre-trained models from other datasets \cite{clothing1m, tcl}, without using any training labels. Although these approaches can mitigate the influence of noisy labels, neglecting label information can result in limited inter-class separability \cite{selfcc}, degrading the performance of label correction. If we can detect clean samples and incorporate these reliable labels into the pre-training of the encoder, the inter-class separability can be improved \cite{simmatch, co-match}. Based on this, we propose an alternating training strategy between noisy meta correction and semi-supervised representation learning to solve \cref{eq:stct_optimization_decompose}, as shown in \cref{fig:diagram_method}. The pseudo code can be referred in Appendix C. Noisy meta correction adjusts the noisy labels based on the embedding features. With these corrected labels, semi-supervised representation learning identifies clean samples and utilizes these detected reliable labels to optimize the distribution of the embedding features. The two steps reinforce each other, forming a positive feedback. A detailed description of the two steps is as follows:

\noindent \textbf{Noisy Meta Correction (NMC)} is responsible for correcting noisy labels. First, a sub-training set $\hat D_{\mathrm{t}}=\{\hat X_{\mathrm{t}},\hat  Y_{\mathrm{t}}\}$ and a noisy validation set $\hat D_{\mathrm{v}}=\{\hat X_{\mathrm{v}}, \hat Y_{\mathrm{v}}\}$ are randomly sampled from the training set $D_{\mathrm{t}}=\{X_{\mathrm{t}}, Y_{\mathrm{t}}\}$ at a sampling rate of $r$. The corresponding embedding features $H_{\mathrm{t}}$ and $H_{\mathrm{v}}$ are extracted with the pre-trained encoder $f_{\mathrm{enc}}$. If the samples are linearly separable in the embedding space, a linear model should be able to perfectly distinguish between each class. Once the linear model performs poorly on the noisy validation set $\hat D_{\mathrm{v}}$, according to Theorems 1 and 2, the source of the poor performance should be the presence of noisy labels in the sub-training set $\hat D_{\mathrm{t}}$.

Since the linear regression model has a closed-form solution, which can greatly simplify the optimization of \cref{eq:stct_optimization_decompose}, we fit a linear regression model on the sub-training set, and $\mathcal{L}_{\text{train}}$ can be expressed as
\begin{equation}
\mathcal{L}_{\text{train}}=\left\| \hat Y _{\text{t}} - H_{\text{t}}\boldsymbol{\omega}\right\|^{2}.
\end{equation}
The predictions on the noisy validation set can be solved via
\begin{equation}
\hat Y'_{\mathrm{v}}=H_{\mathrm{v}}\left( H_{\mathrm{t}}^{\top}H_{\mathrm{t}} \right)^{-1}H_{\mathrm{t}}^{\top}\hat Y_{\mathrm{t}}.
\end{equation}
Despite the accuracy of the optimal model on the noisy distribution is $\sum\limits_{i = 1}^C {\Pr \left[ {Y = i} \right]{T_{i,i}}}$, considering the limited sample size of $\hat D_{\mathrm{v}}$, the meta task is simplified to matching $\hat Y'_{\mathrm{v}}$ with $\hat Y_{\mathrm{v}}$, and MSE is employed to quantify the discrepancy 
\begin{equation}
\mathcal{L}_{\text{val}}=\left\| \hat Y _{\text{v}} - H_{\mathrm{v}}\left( H_{\mathrm{t}}^{\top}H_{\mathrm{t}} \right)^{-1}H_{\mathrm{t}}^{\top}\hat Y_{\mathrm{t}}\right\|^{2}.
\end{equation}
The meta label correction process can be formulated as
\begin{equation}
\small
\hat Y_t \leftarrow \hat Y_t + \frac{2 \eta}{n_{\mathrm{v}}} H_\mathrm{t} (H_\mathrm{t}^{\top} H_\mathrm{t})^{-1} H_\mathrm{v}^{\top} \left(\hat Y_\mathrm{v} - H_\mathrm{v} (H_\mathrm{t}^{\top} H_\mathrm{t})^{-1} H_\mathrm{t}^{\top} \hat Y_t \right),
\end{equation}
where $\eta$ is the learning rate. In Appendix D, a theoretical analysis of label correction is provided.

After the labels in the sub-training set are corrected, we update the corresponding labels in the training set $\{X_\mathrm{t}, Y_\mathrm{t}\}$. Then the updated training data is once again divided into a new sub-training set and a new noisy validation set through random sampling for further label correction. This process is repeated until NMC converges. If the change in corrected training labels between two successive samples falls below a stopping threshold $\delta$, or if the sampling times reaches $\frac{\text{ln}\left( 1-\text{exp}\left( \text{ln}\left( \beta \right)/n \right) \right)}{\text{ln}\left( 1-r \right)}$, NMC is considered to have converged.

\noindent \textbf{Semi-supervised Representation Learning (SRL)} plays the role of improving the inter-class separability of the embedding features and classification. First, the clean samples are identified by measuring the agreement between the embedding features and the corrected training labels. Using the cosine distance to measure the distance between samples in the embedding space, we aggregate the labels of each sample's $k$-nearest neighbors to generate a probabilistic pseudo label $Y_{\mathrm{p}}$. Then the clean samples in each class can be selected via
\begin{equation}
\mathbb{S}_{\mathrm{C}}=\left\{ \left( x_{i},y_{i} \right)|\ell_{\mathrm{CE}}\left(y_{i},y_{\mathrm{p},i}  \right)<\mu_{c}, y_i=c \right\},
\end{equation}
where $\ell_{\mathrm{CE}}\left(.,.\right)$ denotes cross-entropy loss, $\mu_{c}$ represents the cross-entropy value corresponding to the top $\hat \mu$ of the samples in the $c$-th class and $\hat{\mu} = \mu * \text{epoch}$, meaning that as the alternating training between NMC and SRL progresses, more labels can be trusted. After identifying clean data, we discard the labels of the remaining samples and split the training set into labeled data $D_{\mathrm{L}}=\{X_{\mathrm{L}}, Y_{\mathrm{L}}\}$ and unlabeled data $D_{\mathrm{U}}=\{X_{\mathrm{U}}\}$ for semi-supervised learning.

\begin{table*}[]
\small
\centering
\setcounter{table}{0}
\renewcommand{\thetable}{\Roman{table}}
\caption{Testing accuracy (\%) on CIFAR-10 and CIFAR-100 with different symmetric label noise. “Extra Clean Data” denotes an extra clean validation set is needed.}
\begin{tabular}{lccccccccccc}
\noalign{\hrule height 1pt}
\multirow{2}{*}{Method} & \multirow{2}{*}{Category}  & \multirow{2}{*}{\begin{tabular}[c]{@{}c@{}}Extra\\ Clean Data\end{tabular}} & \multicolumn{4}{c}{CIFAR-10} &  & \multicolumn{4}{c}{CIFAR-100} \\ \cline{4-7} \cline{9-12} 
                        &                            &                                                                             & 20\%\rule{0pt}{0.9em}   & 50\%   & 80\%   & 90\%  &  & 20\%   & 50\%   & 80\%   & 90\%   \\ \noalign{\hrule height 1pt}
Cross-Entropy\rule{0pt}{0.9em}           &                            & \(\times\)                                                                            & 82.7  & 57.9  & 26.1  & 16.8 &  & 61.8  & 37.3  & 8.8   & 3.5   \\\hdashline
PENCIL \cite{pencil}\rule{0pt}{0.9em}                  & \multirow{7}{*}{Non-Meta} & \(\times\)                                                                            & 92.4  & 89.1  & 77.5  & 58.9 &  & 69.4  & 57.5  & 31.1  & 15.3  \\
DivideMix \cite{dividemix}\rule{0pt}{0.9em}               &                            & \(\times\)                                                                            & \textbf{96.1}  & 94.6  & 93.2  & 76.0 &  & 77.3  & 74.6  & 60.2  & 31.5  \\
Sel-CL+ \cite{sel-cl}\rule{0pt}{0.9em}                 &                            & \(\times\)                                                                      & 95.5  & 93.9  & 89.2  & 81.9 &  & 76.5  & 72.4  & 59.6  & 48.8  \\
MOIT+ \cite{moit}\rule{0pt}{0.9em}                   &                            & \(\times\)                                                                            & 94.1  & 91.8  & 81.1  & 74.7 &  & 75.9  & 70.6  & 47.6  & 41.8  \\
TCL \cite{tcl}\rule{0pt}{0.9em}                     &                            & \(\times\)                                                                            & 95.0  & 93.9  & 92.5  & 89.4 &  & 78.0  & 73.3  & 65.0  & 54.5  \\
PLM \cite{plm}\rule{0pt}{0.9em}                     &                            & \(\times\)                                                                            & 91.1  & 85.1  & -     & -    &  & 69.5  & 60.4  & -     & -     \\
L2B \cite{l2b}\rule{0pt}{0.9em}                     &                            & \(\times\)                                                                            & \textbf{96.1}  & 95.4  & 94.0  & 91.3 &  & 77.9  & 75.9  & 62.2  & 35.8 \\ \hdashline
MLNT \cite{mlnt}\rule{0pt}{0.9em}                    & \multirow{7}{*}{Meta}      & \(\checkmark\)                                                                             & 92.9  & 89.3  & 77.4  & 58.7 &  & 68.5  & 59.2  & 42.4  & 19.5  \\
MLC \cite{mlc}\rule{0pt}{0.9em}                     &                            & \(\checkmark\)                                                                             & 92.6  & 88.1  & 77.4  & 67.9 &  & 66.8  & 52.7  & 21.8  & 15.0  \\
MSLC \cite{mslc}\rule{0pt}{0.9em}                    &                            & \(\checkmark\)                                                                             & 93.4  & 89.9  & 69.8  & 56.1 &  & 72.5  & 65.4  & 24.3  & 16.7  \\
CTO \cite{cto}\rule{0pt}{0.9em}                     &                            & \(\checkmark\)                                                                             & 87.9  & 84.9  & 80.1  & -    &  & 58.9  & 52.8  & 42.7  & -     \\
DMLP \cite{dmlp}\rule{0pt}{0.9em}                    &                            & \(\checkmark\)                                                                             & 94.7  & 84.2  & 93.5  & 92.8 &  & 72.7  & 68.0  & 63.5  & 61.3  \\
FSR \cite{noisy_val1}\rule{0pt}{0.9em}                    &                            & \(\times\)                                                                            & 95.1  & -  & 82.8  & - &  & \textbf{78.7}  & -  & 46.7  & -  \\
\textbf{STCT}\rule{0pt}{0.9em}                    &                            & \(\times\)                                                                            & 96.0  & \textbf{95.6}  & \textbf{95.2}  & \textbf{95.0} &  & 78.4  & \textbf{76.5}  & \textbf{72.1}  & \textbf{69.3}  \\ \noalign{\hrule height 1pt}
\end{tabular}
\label{tab:cifar_symmetric}
\end{table*}

Following previous works \cite{simmatch, co-match, semi1}, SRL consists of an encoder $f_{\mathrm{enc}}$, a classification head $g_{\mathrm{cls}}$ and a projection head $g_{\mathrm{proj}}$. Firstly, for labeled data, we use weak augmentation functions to obtain augmented samples $X_{\mathrm{L}}^{\mathrm{w}}$. The optimization objective for labeled data can be expressed as
\begin{equation}
\mathcal{L}_{\mathrm{L}} = \frac{1}{n_{\mathrm{L}}}\sum_{i=1}^{n_{\mathrm{L}}}\ell_{\mathrm{CE}}\left( \boldsymbol{p^{\mathrm{w}}_{\mathrm{L},i}}, y_{\mathrm{L},i} \right),
\end{equation}
where $n_{\mathrm{L}}$ is the sample size of the labeled data and $\boldsymbol{p^{\mathrm{w}}_{\mathrm{L},i}}=g_{\mathrm{cls}}(f_{\mathrm{enc}}(\boldsymbol{x^{\mathrm{w}}_{\mathrm{L},i}}))$ denotes the predictive probability of the $i$-th labeled sample. Additionally, weak and strong augmentations are applied to unlabeled data, yielding prediction probabilities $P_{\mathrm{U}}^{\mathrm{w}}$ and $P_{\mathrm{U}}^{\mathrm{s}}$, respectively. The objective function for unlabeled data is
\begin{equation}
\mathcal{L}_{\mathrm{U}} = \frac{1}{n_{\mathrm{U}}}\sum_{i=1}^{n_{\mathrm{U}}}\mathbb{I}_{\mathrm{max}(\boldsymbol{p_{\mathrm{U},i}^{\mathrm{w}}})>\lambda}\ell_{\mathrm{CE}}\left( \boldsymbol{p_{\mathrm{U},i}^{\mathrm{s}}}, \boldsymbol{p_{\mathrm{U},i}^{\mathrm{w}}} \right),
\end{equation}
where $n_{\mathrm{U}}$ is the number of unlabeled data and $\lambda$ is the confidence threshold. Additionally, inspired by \cite{simclr, simmatch, co-match}, the instance-level similarity of unlabeled data is incorporated to regulate the training of SRL. In other words, the spatial relationships between two augmented versions of a sample and other samples are supposed to be similar. To implement this regularization, we first randomly sample $M$ unlabeled samples $X_{\mathrm{Con}}$ and extract their low-dimensional projections with $Z_{\mathrm{Con}}=g_{\mathrm{proj}}(f_{\mathrm{enc}}(X_{\mathrm{Con}}))$. The instance-level similarity between the $i$-th weakly augmented sample in $X_{\mathrm{U}}$ and the $j$-th sample in $X_{\mathrm{Con}}$ can be solved through
\begin{equation}
\small
s_{\text{Con},ij}^{\text{w}}=\frac{\text{exp}\left( \text{sim}\left(\boldsymbol{z_{\text{Con},j}},\boldsymbol{z_{\text{U},i}^{\text{w}}}  \right) /\tau \right)}{\sum_{m=1}^{M}\text{exp}\left( \text{sim}\left(\boldsymbol{z_{\text{Con},m}},\boldsymbol{z_{\text{U},i}^{\text{w}}}  \right) /\tau \right)},
\end{equation}
where $\mathrm{sim}(.,.)$ is the cosine similarity and $\tau$ is the temperature. The strong augmentation's similarity $s_{\text{Con},ij}^{\text{s}}$ can be solved in the same way. The consistency between the strong and weak augmentations can obtained via
\begin{equation}
\mathcal{L}_{\mathrm{Con}} = \frac{1}{n_{\mathrm{U}}}\sum_{i=1}^{n_{\mathrm{U}}}\ell_{\mathrm{CE}}\left(\boldsymbol{s_{\text{Con},i}^{\text{s}}},\boldsymbol{s_{\text{Con},i}^{\text{w}}}\right).
\end{equation}
Finally, the objective function of SRL is
\begin{equation}
\mathcal{L}_{\mathrm{SRL}} = \mathcal{L}_{\mathrm{L}} + \mathcal{L}_{\mathrm{U}} + \mathcal{L}_{\mathrm{Con}}.
\end{equation}
After training SRL, the well-trained encoder is employed to extract embedding features in NMC step. It should be mentioned that before conducting noisy meta correction for the first time, we follow previous studies \cite{dmlp, tcl} and label information is not used to pre-train the encoder, $\theta^{\ast}\gets F_{\mathrm{Pretrain}}(X_{\mathrm{t}})$. Once the training labels are corrected, the encoder is then pre-trained with SRL, $\theta^{\ast} \gets F_{\mathrm{Pretrain}}(D_{\mathrm{L}}, D_{\mathrm{U}})$. Finally, the testing results can be predicted via $P_{\mathrm{test}}=g_{\mathrm{cls}}(f_{\mathrm{enc}}(X_{\mathrm{test}}))$.

The pseudo code of solving STCT is summarized as following:

\begin{algorithm}
\caption{Pseudo Code of STCT.}
\label{alg1}
\begin{algorithmic}
\STATE \textbf{Input:} Training set $D_{\mathrm{t}}=\{X_{\mathrm{t}},Y_{\mathrm{t}}\}$; Encoder $f_{\mathrm{enc}}$; Classification head $g_{\mathrm{cls}}$; Projection head $g_{\mathrm{proj}}$.
\STATE \textbf{Output:} Well-trained $f_{\mathrm{enc}}$ and $g_{\mathrm{cls}}$.
\vspace{2.5mm}
\STATE Initialize $f_{\mathrm{enc}}$ with $F_{\mathrm{Pretrain}}(X_{\mathrm{t}})$;
\FOR{$\mathrm{epoch}$ in $\{1,2,...,\mathrm{max\_epoch}\}$}
    \vspace{1mm}
    \STATE // Noisy Meta Correction (NMC)
    \vspace{1mm}
    \WHILE{True}
        \STATE Divide $D_{\mathrm{t}}$ into sub-training set $\hat D_{\mathrm{t}}$ and noisy validation set $\hat D_{\mathrm{v}}$ through random sampling;
        \STATE Extract embedding features $H_{\mathrm{t}}$ and $H_{\mathrm{v}}$ with $f_{\mathrm{enc}}$;
        \STATE Fit a linear regression model with Eq. (6);
        \STATE Correct noisy labels in $\hat D_{\mathrm{t}}$ with Eqns. (7-9);
        \STATE Update the corrected labels to $D_{\mathrm{t}}$;
        \IF{one of the early stopping criteria is met}
            \STATE break;
        \ENDIF
    \ENDWHILE
    \vspace{1mm}
    \STATE // Self-supervised Representation Learning (SRL)
    \vspace{1mm}
    \STATE Select clean samples with Eq. (10);
    \STATE Optimize $f_{\mathrm{enc}}$, $g_{\mathrm{cls}}$ and $g_{\mathrm{proj}}$ with Eq. (15);
\ENDFOR
\STATE \textbf{Return} Well-trained $f_{\mathrm{enc}}$ and $g_{\mathrm{cls}}$.
\end{algorithmic}
\end{algorithm}

\section{Experiments}
\noindent In this section, extensive experiments are conducted on various synthetic and real-world noisy datasets to verify the effectiveness of the proposed STCT.

\subsection{Experimental Settings}

\textbf{Synthetic Datasets} \, We evaluate the performance of STCT on CIFAR-10 and CIFAR-100. Both datasets contain 50,000 training images and 10,000 testing images. Two types of synthetic label noise, symmetric and asymmetric, are introduced to contaminate the training data. Symmetric noise refers to modifying training labels to arbitrary labels according to predefined noise rates, while asymmetric noise, on the other hand, involves flipping training labels to semantically related categories. The flipping relations for asymmetric noise follow the settings in \cite{ctrr, mlnt}. The noise rates are set to 20\%, 50\%, 80\% and 90\% for symmetric noise, and 20\% and 40\% for asymmetric noise.

For parameter settings of synthetic datasets, we use WRN28-2 and WRN28-8 as the encoder for CIFAR-10 and CIFAR-100, respectively \cite{wideresnet}. The encoder is initialized with SimCLR \cite{simclr, dpac}. The hyper-parameters are set as $r=0.5$, $\delta=0.998$, $\eta=20\%$, $\lambda=0.95$, $\tau=0.1$, $\beta=0.9999$. Considering the sample size in each class, $k$ is set to 2000 for CIFAR-10 and 300 for CIFAR-100. SGD is chosen to be the optimizer whose learning rate is 0.002 and momentum is 0.9.

\noindent\textbf{Real-world Datasets} The capability of STCT to combat real-world label noise is evaluated on CIFAR-10N, CIFAR-100N and Clothing-1M. The image data in CIFAR-10N/100N is the same as that in CIFAR-10/100, while the label data contains real-world human annotation errors \cite{cifar10n_dataset}.
Clothing-1M is a large-scale noisy image dataset collected from several online shopping websites, with only the noisy training set being used in this study \cite{clothing1m}.

Most hyper-parameter settings of STCT on the real-world datasets are identical to those used for the synthetic datasets, except for Clothing-1M, where, following the settings in \cite{dividemix, ot-filter, sample-wise}, an ImageNet-pretrained ResNet-50 is used to initialize the encoder, $\lambda=0.7$ and $k=2000$.

\begin{figure}[]
    \centering
    \includegraphics[width=0.43\textwidth]{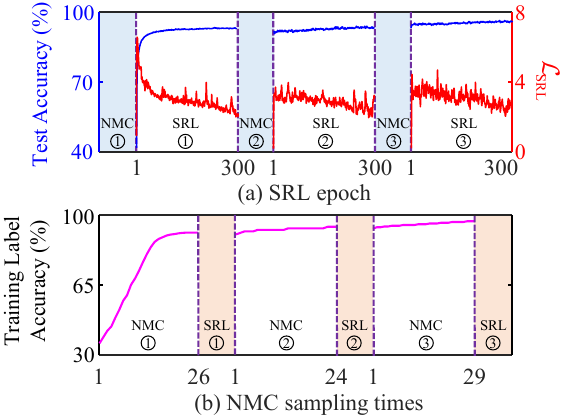}
    \caption{The training process of STCT on CIFAR-10 with 80\% symmetric noise. (a) The testing accuracy and training loss curves of SRL. (b) The corrected training label accuracy curve of NMC. }
    \label{fig:NMC_SRL}
\end{figure}

\begin{table}[]
\small
\centering
\setcounter{table}{1}
\renewcommand{\thetable}{\Roman{table}}
\caption{Testing accuracy (\%) on CIFAR-10 and CIFAR-100 with different asymmetric label noise.}
\begin{tabular}{lccccc}
\noalign{\hrule height 1pt}
\multirow{2}{*}{Method} & \multicolumn{2}{c}{CIFAR-10}   \rule{0pt}{0.9em}                     &                      & \multicolumn{2}{c}{CIFAR-100}                       \\ \cline{2-3} \cline{5-6} 
                        & 20\%  \rule{0pt}{0.9em}                  & 40\%                     &                      & 20\%                     & 40\%                     \\ \noalign{\hrule height 1pt}
Cross-Entropy \rule{0pt}{0.9em}          & 81.3                     & 77.1                     &                      & 61.3                     & 44.5                     \\
PENCIL \rule{0pt}{0.9em}                 & 92.4                     & 91.2                     &                      & 74.7                     & 63.6                     \\
Sel-CL+ \rule{0pt}{0.9em}                & 95.2                     & 93.4                     &                      & \textbf{77.5}            & 74.2                     \\
DivideMix \rule{0pt}{0.9em}                    & 93.4                     & 91.7                     &                      & 76.2                        & 55.1                        \\
TCL \rule{0pt}{0.9em}                    & 94.6                     & 93.7                     &                      & -                        & -                        \\
MLNT \rule{0pt}{0.9em}                   & -                        & 90.2                     &                      & -                        & -                        \\
MSLC \rule{0pt}{0.9em}                   & 94.4                     & 91.6                     &                      & 72.7                     & 69.2                     \\
DMLP \rule{0pt}{0.9em}                   & 94.6                     & 93.9                     &                      & 73.3                     & 68.4                     \\
\textbf{STCT} \rule{0pt}{0.9em}                   & \textbf{95.9}            & \textbf{95.6}            &                      & 76.8                     & \textbf{75.6}            \\ \noalign{\hrule height 1pt}
\end{tabular}

\label{tab:cifar_asymmetric}
\end{table}

\begin{table*}[!t]
\small
\centering
\setcounter{table}{3}
\renewcommand{\thetable}{\Roman{table}}
\caption{Testing accuracy (\%) on CIFAR-10N and CIFAR-100N.}
\begin{tabular}{lccccccc}
\noalign{\hrule height 1pt}
\multirow{2}{*}{Method} & \multicolumn{5}{c}{CIFAR-10N}                                                 &           & CIFAR-100N    \\ \cline{2-6} \cline{8-8} 
                        & Aggregate\rule{0pt}{0.9em}     & Random1       & Random2       & Random3       & Worst         &           & Noisy100      \\ \noalign{\hrule height 1pt}
Cross-Entropy\rule{0pt}{0.9em}           & 87.8          & 85.0          & 86.5          & 85.2          & 77.7          &           & 55.5          \\
CORES$^2$ \cite{cores} \rule{0pt}{0.9em}                   & 95.3          & 94.5          & 94.9          & 94.7          & 91.7          &           & 64.2          \\
GNL \cite{gnl}\rule{0pt}{0.9em}                     & 95.3          & 92.0          & 91.4          & 91.8          & 87.0          &           & 64.5          \\
DivideMix \cite{dividemix}\rule{0pt}{0.9em}               & 95.6          & 95.3          & 95.3          & 95.4          & 93.2          &           & 71.1          \\
ELR+ \cite{elr}\rule{0pt}{0.9em}                     & 94.8          & 94.4          & 94.2          & 94.3          & 91.1          &           & 66.7          \\
SOP \cite{sop}\rule{0pt}{0.9em}                     & 95.6          & 95.3          & 95.3          & 95.4          & 93.2          &           & 67.8          \\
SGN \cite{sgn}\rule{0pt}{0.9em}                     & 92.1          & 91.9          & 91.7          & 91.9          & 86.7         &           & 60.4          \\
\textbf{STCT}           & \textbf{95.7} & \textbf{95.8} & \textbf{95.8} & \textbf{95.7} & \textbf{95.4} & \textbf{} & \textbf{73.6} \\ \noalign{\hrule height 1pt}
\end{tabular}
\label{tab:acc-cifar10n-100n}
\end{table*}

\begin{table}[]
\centering
\small
\setcounter{table}{2}
\renewcommand{\thetable}{\Roman{table}}
\caption{The corrected training label accuracy (\%) of label correction methods on CIFAR-10 and CIFAR-100. “S” and “A” represent symmetric and asymmetric noise, respectively.}
\begin{tabular}{clccc}
\noalign{\hrule height 1pt}
\rule{0pt}{1em}Dataset                    & Method        & S-50\%        & S-80\%        & A-40\%        \\ \noalign{\hrule height 1pt}
\multirow{6}{*}{CIFAR-10}  & \rule{0pt}{1em}PENCIL        & 90.2          & 68.3          & 79.8          \\
                           & \rule{0pt}{1em}DivideMix     & 93.7             & 88.5             & 42.5             \\
                           & \rule{0pt}{1em}MLC           & 73.1          & 38.6          & 55.6             \\
                           & \rule{0pt}{1em}MSLC          & 87.2          & 68.5          & 20.1             \\
                           & \rule{0pt}{1em}DMLP          & 92.1          & 86.3          & 86.1          \\
                           & \rule{0pt}{1em}\textbf{STCT} & \textbf{97.4} & \textbf{96.9} & \textbf{97.9} \\ \noalign{\hrule height 1pt}
\multirow{6}{*}{CIFAR-100} & \rule{0pt}{1em}PENCIL        & 70.2          & 48.1          & 61.5          \\
                           & \rule{0pt}{1em}DivideMix     & 78.9             & 56.9             & 63.2             \\
                           & \rule{0pt}{1em}MLC           & 15.0             & 13.4             & 14.2             \\
                           & \rule{0pt}{1em}MSLC          & 64.5          & 20.8         & 46.9          \\
                           & \rule{0pt}{1em}DMLP          & 81.3          & 65.2          & 68.7          \\
                           & \rule{0pt}{1em}\textbf{STCT} & \textbf{85.2} & \textbf{73.5}    & \textbf{71.9}    \\ \noalign{\hrule height 1pt}
\end{tabular}

\label{tab:corrected_label}
\end{table}

\subsection{Model Performance}

\textbf{Synthetic Datasets} The testing accuracy of the proposed STCT and other competing methods on CIFAR-10 and CIFAR-100 with different label noise is reported in \cref{tab:cifar_symmetric} and \cref{tab:cifar_asymmetric}. In this study, the performance of the competing methods is collected from their reports and \cite{dmlp, ot-filter, tcl, sop}. According to the experimental results, STCT achieves the best performance in most cases, especially for high noise rate. For instance, under 90\% symmetric noise, STCT achieves 2.2\% and 8.0\% performance improvements compared to the second best method on CIFAR-10 and CIFAR-100, respectively. Besides, we divide all the competing methods into meta-learning and non-meta learning based methods. Most meta learning methods require an extra clean validation set, whereas STCT outperforms these methods without needing any additional data. What's more, in FSR and FaMUS \cite{noisy_val2}, a clean validation set can be constructed through pseudo-label and sample selection and also do not require additional clean data. (Notably, since in FaMUS, only the results on CIFAR-10/100 with 40\% and 60\% symmetric noise are reported, given the different noise conditions, we do not present the results in \cref{tab:cifar_symmetric}.) According to the performance reported in \cite{noisy_val2}, on CIFAR-10, STCT surpasses FaMUS for 0.5\% and 0.7\% under 40\% and 60\% symmetric noise. For CIFAR-100, the improvements of STCT are 1.3\% and 2.4\%, respectively. The better performance of STCT compared with FSR and FaMUS suggests that, although these two meta learning methods do not need extra data, the clean validation set selected based on manually designed rules may underperform compared to randomly sampled noisy validation sets due to biased estimation of the clean distribution.

We attribute the good performance of STCT on classification to its powerful label correction capability. In \cref{tab:corrected_label}, the corrected training label accuracy of STCT and other label correction methods is reported. The label correction results indicate that STCT greatly outperforms the competing methods, particularly under high noise rate conditions. Specifically, STCT achieves a training label accuracy of 96.9\% on CIFAR-10 and 73.5\% on CIFAR-100 with 80\% symmetric noise, achieving improvements of 8.4\% and 8.3\%.

\noindent\textbf{Real-world Datasets} We evaluate the performance of STCT on three real-world datasets: CIFAR-10N, CIFAR-100N, and Clothing-1M, with the results presented in \cref{tab:acc-cifar10n-100n} and \cref{tab:clothing-1m}. We can find that STCT exhibits excellent performance under real-world label noise. STCT achieves a 2.2\% performance improvement over SOP and DivideMix on CIFAR-10N with the Worst label noise, validating its ability to handle real-world noisy labels with high noise rates. Additionally, the 76.0\% testing accuracy achieved by STCT on Clothing-1M demonstrates its effectiveness on large-scale datasets.

\begin{table}[]
\small
\centering
\setcounter{table}{4}
\renewcommand{\thetable}{\Roman{table}}
\caption{Testing accuracy (\%) on Clothing-1M.}
\begin{tabular}{lc:lc}
\noalign{\hrule height 1pt}
Method        & Accuracy & Method        & Accuracy      \\ \noalign{\hrule height 1pt}
Cross-Entropy & 70.9     & FaMUS \cite{famus}         & 74.4          \\
PENCIL \cite{pencil}        & 73.5     & RRL \cite{rrl}         & 74.5         \\
C2D \cite{c2d}           & 74.3     & CORES$^2$  \cite{cores}       & 73.2          \\
GCE \cite{gce}           & 73.3     & DivideMix \cite{dividemix}    & 74.8          \\
MLNT \cite{mlnt}          & 73.5     & CleanNet  \cite{cleannet}    & 74.7          \\
MLC \cite{mlc}          & 75.8     & PLC \cite{plc}          & 74.0          \\
MSLC  \cite{mslc}        & 74.0     & \textbf{STCT} & \textbf{76.0} \\ \noalign{\hrule height 1pt}
\end{tabular}

\label{tab:clothing-1m}
\end{table}

\subsection{Ablation Studies}

\textbf{Contribution of different training steps} \, STCT consists of two alternating training steps: noisy meta correction (NMC) and semi-supervised representation learning (SRL).  We explore the contribution of NMC and SRL on the classification and label correction performance of STCT in \cref{tab:ablation_NMC_SRL}. Without NMC, the noisy labels cannot be corrected and the clean samples are selected based on the original training data. As a result, the selected clean samples are inaccurate and limited in number, resulting in a 3.1\% decrease in classification accuracy on CIFAR-10 with 80\% symmetric noise. When SRL is offline, the corrected training labels are used to train a classifier solely with cross-entropy loss. Although the training labels can be corrected, the encoder is pre-trained without using any label information, resulting in limited linear separability between classes. This limitation reduces the accuracy of the training labels and subsequently affects classification performance. The impact of SRL on STCT's performance is particularly evident on CIFAR-100, where the training label accuracy decreases by 14.1\% and 13.3\% under 50\% and 80\% symmetric noise, respectively. Furthermore, we present the training process of STCT on CIFAR-10 with 80\% symmetric noise in \cref{fig:NMC_SRL}, visually demonstrating the mutual reinforcement between NMC and SRL.

\noindent\textbf{Contribution of different losses in SRL} SRL is responsible for classification and optimizing embedding features, which is critical to the performance of STCT. In \cref{tab:ablation_losses}, we report the contributions of ${{\cal L}_\mathrm{L}}$, ${{\cal L}_\mathrm{U}}$ and ${{\cal L}_\mathrm{Con}}$ to STCT's performance. Experimental results indicate that when the three losses are offline, both the classification and label correction performance decline. This finding demonstrates that supervised learning on clean samples ${{\cal L}_\mathrm{L}}$, utilization of unlabeled data's pseudo-labels ${{\cal L}_\mathrm{U}}$, and constraints on instance similarity ${{\cal L}_\mathrm{Con}}$ collectively enhance the separability of embedding features, thereby improving STCT's performance.

\begin{table}[]
\centering
\caption{Ablation study for the contribution of NMC and SRL on CIFAR-10 and CIFAR-100 with 50\% and 80\% symmetric noise.}
\small
\begin{tabular}{clccccc}
\noalign{\hrule height 1pt}
\multicolumn{1}{l}{\multirow{2}{*}{}}                                                 & \rule{0pt}{0.9em}\multirow{2}{*}{Method} & \multicolumn{2}{c}{CIFAR-10}  &           & \multicolumn{2}{c}{CIFAR-100} \\ \cline{3-4} \cline{6-7} 
\multicolumn{1}{l}{}                                                                  &                   & \rule{0pt}{0.9em}50\%          & 80\%          &           & 50\%          & 80\%          \\ \noalign{\hrule height 1pt}
\multirow{3}{*}{\begin{tabular}[c]{@{}c@{}}Testing\\ Accuracy\end{tabular}}           & \rule{0pt}{0.9em}w/o NMC           & 93.6          & 92.1          &           & 70.1          & 63.1          \\
                                                                                      & \rule{0pt}{0.9em}w/o SRL           & 92.1          & 91.3          &           & 67.6          & 61.5         \\
                                                                                      & \rule{0pt}{0.9em}\textbf{STCT}     & \textbf{95.6} & \textbf{95.2} & \textbf{} & \textbf{76.5} & \textbf{72.1} \\ \noalign{\hrule height 1pt}
\multirow{3}{*}{\begin{tabular}[c]{@{}c@{}}Training \\ Label\\ Accuracy\end{tabular}} & \rule{0pt}{0.9em}w/o NMC           & -          & -          &           & -          & -          \\
                                                                                      & \rule{0pt}{0.9em}w/o SRL           & 91.9             & 90.3             &           & 71.1             & 60.2             \\
                                                                                      & \rule{0pt}{0.9em}\textbf{STCT}     & \textbf{97.4} & \textbf{96.9} & \textbf{} & \textbf{85.2} & \textbf{73.5}    \\ \noalign{\hrule height 1pt}
\end{tabular}
\label{tab:ablation_NMC_SRL}
\end{table}

\begin{table}[]
\centering
\caption{Ablation study for the contribution of ${{\cal L}_\mathrm{L}}$, ${{\cal L}_\mathrm{U}}$ and ${{\cal L}_\mathrm{Con}}$ on CIFAR-10 and CIFAR-100 with 50\% and 80\% symmetric noise. “N-C” indicates that the algorithm does not converge.}
\small
\begin{tabular}{clccccc}
\noalign{\hrule height 1pt}
\multicolumn{1}{l}{\multirow{2}{*}{}}                                                 & \rule{0pt}{0.9em}\multirow{2}{*}{Method} & \multicolumn{2}{c}{CIFAR-10}  &           & \multicolumn{2}{c}{CIFAR-100} \\ \cline{3-4} \cline{6-7} 
\multicolumn{1}{l}{}                                                                  &                   & \rule{0pt}{0.9em}50\%          & 80\%          &           & 50\%          & 80\%          \\ \noalign{\hrule height 1pt}
\multirow{4}{*}{\begin{tabular}[c]{@{}c@{}}Testing\\ Accuracy\end{tabular}}           & \rule{0pt}{0.9em}w/o ${{\cal L}_\mathrm{L}}$           & N-C          & N-C           &           & N-C           & N-C           \\
& \rule{0pt}{0.9em}w/o ${{\cal L}_\mathrm{U}}$           & 93.8          & 93.6          &           & 67.6          & 61.5          \\
                                                                                      & \rule{0pt}{0.9em}w/o ${{\cal L}_\mathrm{Con}}$           & 83.0          & 82.1          &           & 70.1          & 63.1          \\
                                                                                      & \rule{0pt}{0.9em}\textbf{STCT}     & \textbf{95.6} & \textbf{95.2} & \textbf{} & \textbf{76.5} & \textbf{72.1} \\ \noalign{\hrule height 1pt}
\multirow{4}{*}{\begin{tabular}[c]{@{}c@{}}Training \\ Label\\ Accuracy\end{tabular}} & \rule{0pt}{0.9em}w/o ${{\cal L}_\mathrm{L}}$           & N-C          & N-C          &           & N-C          & N-C          \\
& \rule{0pt}{0.9em}w/o ${{\cal L}_\mathrm{U}}$           & 94.8          & 93.7          &           & 71.1          & 60.2          \\
                                                                                      & \rule{0pt}{0.9em}w/o ${{\cal L}_\mathrm{Con}}$           & 83.4             & 81.6             &           & 73.6             & 65.5             \\
                                                                                      & \rule{0pt}{0.9em}\textbf{STCT}     & \textbf{97.4} & \textbf{96.9} & \textbf{} & \textbf{85.2} & \textbf{73.5}    \\ \noalign{\hrule height 1pt}
\end{tabular}
\label{tab:ablation_losses}
\end{table}

\subsection{Limitations of Manually Designed Validation Set}

\begin{figure*}[]
    \centering
    \includegraphics[width=0.7\textwidth]{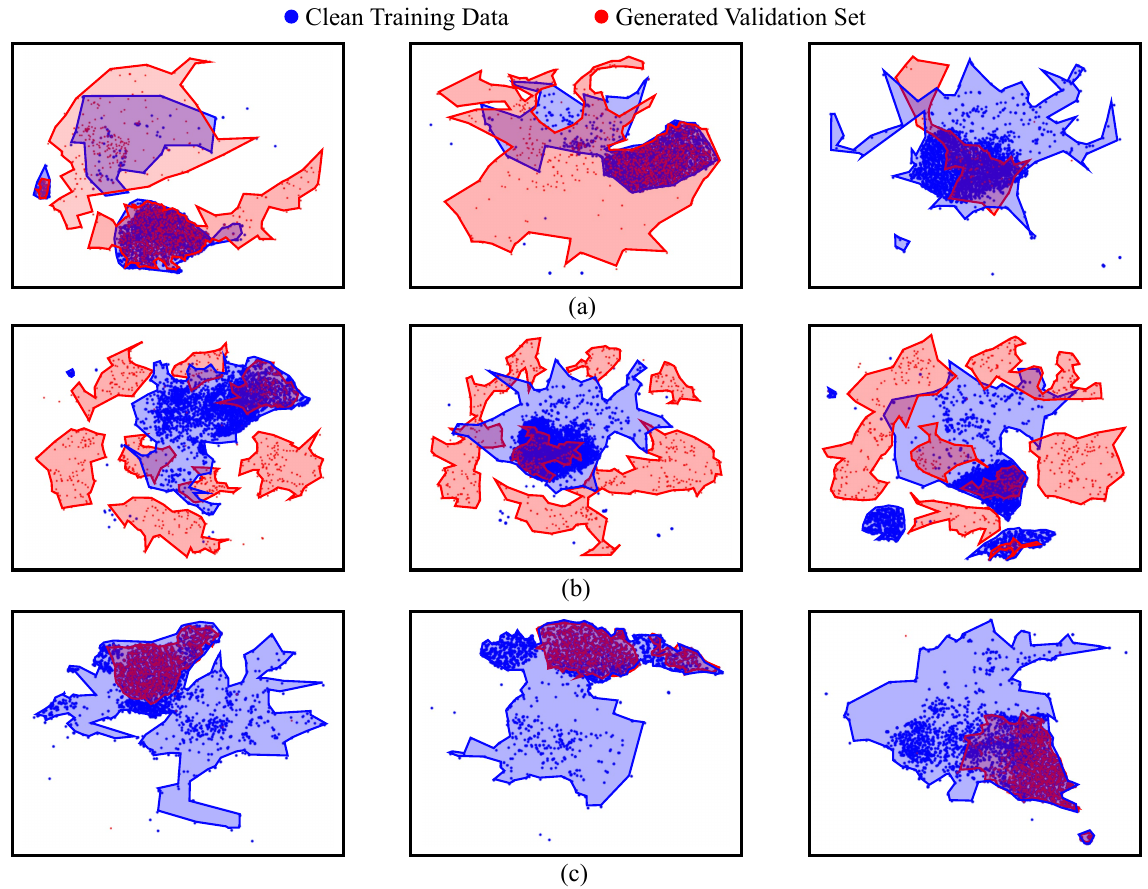}
    \caption{Visualization of the distribution of different generated validation sets and clean data in the embedding space under 50\% symmetric label noise. The blue region represents the clean distribution, and the red region represents the distribution estimated by the generated validation sets. The closer the two regions are, the more effective the validation set is. (a) Loss-based method; (b) KNN-based method; (c) Clustering-based method.}
    \label{fig:distribution_comp}
\end{figure*}

Although most meta learning based LNL methods rely on an extra clean validation sets \cite{dmlp, cto, mlc, mslc, mlnt, meta_weight1, meta_weight2, meta_weight3}, we do notice that some approaches attempt to generate a ``clean'' validation set through sample selection and pseudo-labeling techniques to avoid introducing extra annotation costs, exemplified by FSR \cite{noisy_val1} and FaMUS \cite{famus}. Although the experimental results on CIFAR-10 and CIFAR-100 have demonstrated that STCT outperforms FSR and FaMUS, it should be noted that these two methods belong to reweighting methods \cite{lnl_survey}, leading to the limitations of simply using model performance to compare the effectiveness of the noisy and generated validation sets. Therefore, we compare the label correction performance of randomly selected noisy validation sets (STCT) and ``clean'' validation sets from different generation methods. The comparison is conducted within the framework of non-nested meta label correction. In other words, we use the generated validation data to replace the noisy validation sets in NMC in the experiments of the comparative methods. Based on the previous studies \cite{famus, noisy_val1, meta_learning_survey}, three validation set generation methods are compared with noisy validation sets we used in STCT, namely:

\noindent \textbf{1) Loss-based method (Loss):} This approach is based on the small loss assumption \cite{dividemix} and models the training loss with a Gaussian distribution, distinguishing clean data from noisy data and using the selected clean samples as the validation set \cite{famus}.

\noindent \textbf{2) KNN-based method (KNN):} Following \cite{sel-cl}, this method aggregates the labels of each sample’s $k$-nearest neighbors in the embedding space to generate a probabilistic pseudo label and selects clean data as the validation set by measuring the discrepancy between the pseudo labels and the training labels.

\noindent \textbf{3) Clustering-based method (Clustering):} The clustering-based method leverages the distribution characteristics of the training data in the embedding space to generate pseudo-labels without using any label information \cite{selfcc}. Specifically, we use fuzzy $c$-means to generate soft cluster assignments and select high confidence samples from these assignments as the validation set.

\begin{table}[]
\centering
\caption{The corrected training label accuracy (\%) for different generated validation sets (Loss, Clustering, and KNN) and the noisy validation set (STCT) on CIFAR-10. “S” and “A” represent symmetric and asymmetric noise, respectively. ``\(\times\)'' indicates the noisy labels cannot be corrected at all.}
\begin{tabular}{lccccc}
\noalign{\hrule height 1pt}
Method        & S-50\%         & S-80\%         &           & A-20\%         & A-40\%         \\ \noalign{\hrule height 1pt}
Loss           & 92.0          & 74.4          &           & 91.9          & \(\times\)              \\
Clustering    & 89.8          & 84.3          &           & 93.7          & 90.1          \\
KNN           & 93.3          & \(\times\)              &           & 93.7          & 82.7          \\
\textbf{STCT} & \textbf{94.3} & \textbf{92.6} & \textbf{} & \textbf{95.5} & \textbf{93.0} \\ \noalign{\hrule height 1pt}
\end{tabular}
\label{tab:comp_gen_val_set}
\end{table}
\noindent We report the comparison results in \cref{tab:comp_gen_val_set}. The embedding features used in NMC are extracted from a SimCLR pre-trained encoder, and the detailed settings for each of the three methods can be found in \cite{dividemix, selfcc, sel-cl}, respectively. The experimental results demonstrate that the noisy validation set outperforms the validation set generated based on manually designed rules across various label noise. Loss-based and KNN-based methods even fail to achieve label correction under 80\% symmetric noise and 40\% asymmetric noise. The poor performance of these three generated validation sets is speculated to arise from the biased estimation of the clean distribution caused by the manually designed rules. To further verify this hypothesis, we visualize the distribution of the generated validation set and clean data in the embedding space using $t$-SNE, as shown in \cref{fig:distribution_comp}. The red points represent the generated validation set, while the blue points denote the clean data. If the generated validation set can effectively evaluate the model's performance on the clean distribution, the red and blue regions should be approximately identical. However, as observed, none of the three methods can provide an accurate estimation of the clean distribution. The above results demonstrate that despite a ``clean'' validation set can be generated through manually designed rules, it leads to a biased estimation of the clean distribution, limiting the model's performance. On the other hand, the noisy validation set, selected via random sampling, can effectively estimate the noisy distribution and, in turn, evaluate the model's performance on the clean distribution, without the biased estimation introduced by rule design.

\subsection{Limitations of Manually Designed Validation Set}

\begin{figure}[]
    \centering
    \includegraphics[width=0.36\textwidth]{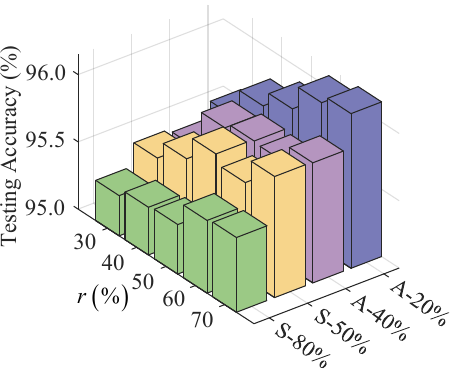}
    \caption{The variation of testing accuracy with different sampling rates $r (\%)$.}
    \label{fig:test_acc_vs_sample_size}
\end{figure}

\begin{figure}[]
    \centering
    \includegraphics[width=0.4\textwidth]{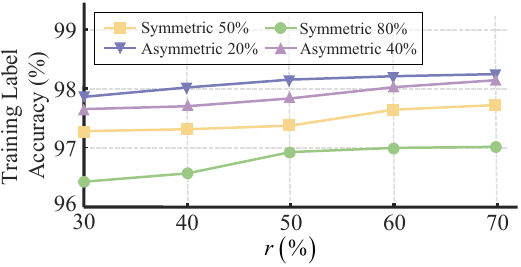}
    \caption{The variation of training label accuracy with different sampling rates $r (\%)$.}
    \label{fig:train_label_acc_vs_sample_size}
\end{figure}

\begin{figure}[]
    \centering
    \includegraphics[width=0.4\textwidth]{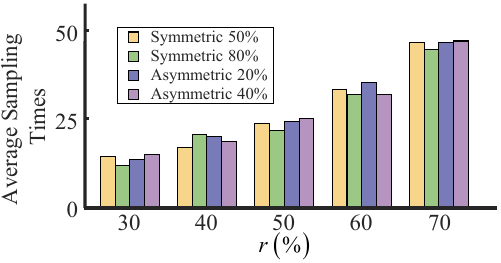}
    \caption{The variation of average sampling times with different sampling rates $r (\%)$.}
    \label{fig:sampling_times_vs_sample_size}
\end{figure}

During each label correction process in NMC, a noisy validation set is randomly sampled from the training set to evaluate model performance on the clean distribution so as to correct noisy labels. According to \cite{theory1}, the size of the sampled noisy validation set affects the estimation of the noisy distribution, thereby impacting the label correction performance. As a result, we investigate the effect of the sampling rate $r$ of the noisy validation set on STCT's performance. In \cref{fig:test_acc_vs_sample_size} and \cref{fig:train_label_acc_vs_sample_size}, we report the testing accuracy and the training label accuracy on CIFAR-10 with different sampling rates. Based on these results, it can be observed that increasing the sampling rate can improve model performance on both classification and label correction. However, an increase in the sample size of the noisy validation set results in a decrease in the number of sub-training samples, leading to a larger sampling time, as shown in \cref{fig:sampling_times_vs_sample_size}. Besides, we also notice that when $r$ is too large, such as $r>90\%$, NMC fails to correct noisy labels. This is because, when the sub-training set's sample size is too small, the model's poor performance primarily stems from a lack of training data, rather than noisy labels. Even with a large noisy validation set, the noisy labels cannot be corrected. Therefore, we recommend setting $r$ to 50\%.

\subsection{Performance Comparison between Noisy Val Sets and Extra Clean Val Sets of Different Sizes}

\begin{figure}[]
    \centering
    \includegraphics[width=0.39\textwidth]{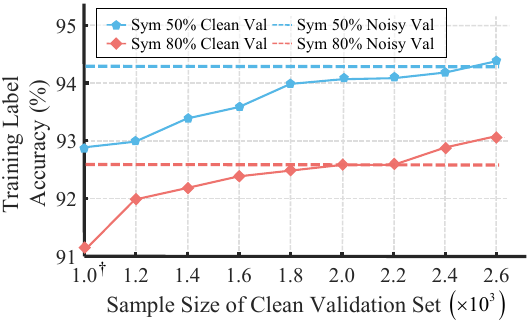}
    \caption{Comparisons between noisy validation sets and extra clean validation sets of different sample sizes on CIFAR-10 with symmetric label noise.}
    \label{fig:clean_val_vs_noisy_val_sym}
\end{figure}

\begin{figure}[]
    \centering
    \includegraphics[width=0.4\textwidth]{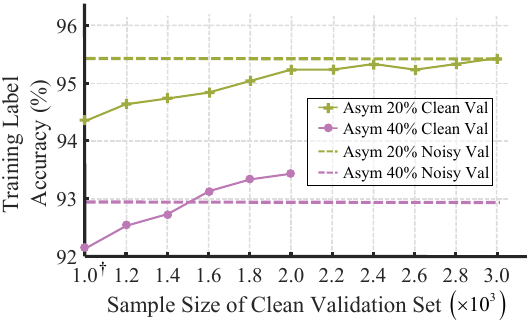}
    \caption{Comparisons between noisy validation sets and extra clean validation sets of different sample sizes on CIFAR-10 with asymmetric label noise.}
    \label{fig:clean_val_vs_noisy_val_asym}
\end{figure}

Experiments are conducted to compare the label correction performance of noisy validation sets with extra clean validation sets of different sample sizes on CIFAR-10. To ensure a fair comparison, the performance of all validation sets is evaluated using the non-nested framework of NMC, with a SimCLR pre-trained encoder to extract the embedding features \cite{simclr, dpac}. The results are reported in \cref{fig:clean_val_vs_noisy_val_sym} and \cref{fig:clean_val_vs_noisy_val_asym}. In previous meta learning based LNL methods \cite{dmlp, famus, noisy_val1, mslc, mlc, mlnt, cto}, the default sample size of the extra clean validation set is 1,000 on CIFAR-10 (denoted by ``$\dagger$'' in \cref{fig:clean_val_vs_noisy_val_sym} and \cref{fig:clean_val_vs_noisy_val_asym}). According to the experimental results, it can be observed that using noisy data as the validation set consistently outperforms the default-sized extra clean validation set in label correction across different label noise. As the size of the clean validation set increases, its correction performance improves gradually and ultimately surpasses that of the noisy validation set. The noisy validation sets exhibit comparable performance to an extra clean validation set containing about 1500-3,000 samples, demonstrating the effectiveness of using noisy data to evaluate model performance on the clean distribution.

\section{Conclusion}
In this paper, we introduce STCT, a novel noisy meta label correction framework for learning from noisy labels. STCT utilizes noisy data to evaluate model performance on clean distribution, achieving label correction through meta learning and eliminating the reliance on extra clean data. Extensive experiments on both synthetic and real-world datasets demonstrate that STCT exhibits superior label correction and classification performance. STCT provides a novel direction for meta learning based LNL methods, making clean data no longer a necessity.

\appendices
\section{}
The proof of Theorem 1 begins by defining the risk of a classifier $f$ on the noisy distribution $\tilde{D}_{\mathrm{t}}$ and the corresponding clean distribution $\tilde{D}_{\mathrm{c}}$,  which are expressed as follows:
\begin{equation}
\small
\mathcal{R}_{\tilde{D}_{\mathrm{t}}}\left( f \right)=\mathbb{E}_{\left( X,Y_{\mathrm{t}} \right)\sim \tilde{D}_{\mathrm{t}}}\left( f\left( X \right)\neq Y_{\mathrm{t}} \right),
\label{eq:risk_D_t}
\end{equation}
\begin{equation}
\small
\mathcal{R}_{\tilde{D}_{\mathrm{c}}}\left( f \right)=\mathbb{E}_{\left( X,Y_{\mathrm{c}} \right)\sim \tilde{D}_{\mathrm{c}}}\left( f\left( X \right)\neq Y_{\mathrm{c}} \right),
\label{eq:risk_D_c}
\end{equation}
where $Y_{\mathrm{t}}$ and  $Y_{\mathrm{c}}$ are the noisy and clean labels, respectively. The noise transition matrix $T$ from $\tilde{D}_{\mathrm{c}}$ to $\tilde{D}_{\mathrm{t}}$ is defined as $T_{ij}=\mathrm{Pr}(Y_{\mathrm{t}}=i|Y_{\mathrm{c}}=j)$, satisfying for $\forall i,j\in \mathcal{Y},0\leqslant T_{ij}\leqslant 1,\sum_{j}^{}T_{ij}=1$. $\mathcal{Y}$ represents the label space.

First, we expand the risk of the classifier $f$ on the noisy distribution. By defining an indicator function $\mathbb{I}(\cdot)$, \cref{eq:risk_D_t} can be reformulated as
\begin{equation}
\small
\mathcal{R}_{\tilde{D}_{\mathrm{t}}}\left( f \right)=\mathbb{E}_{X\sim \tilde{D}_{\mathrm{t}}}\left[\sum_{j\in \mathcal{Y}}^{}\text{Pr}\left(Y_{\mathrm{t}}=j|X \right) \mathbb{I} \left(f\left( X \right)\neq j \right)\right].
\label{eq:risk_D_t_redefined}
\end{equation}
Considering $\text{Pr}\left(Y_{\mathrm{t}}=j|X \right)$ in  \cref{eq:risk_D_t_redefined}, the clean label $Y_{\mathrm{c}}$ is introduced as an intermediate variable as follows:
\begin{equation}
\begin{split}
\small
\text{Pr}\left(Y_{\mathrm{t}}=j|X \right) &  \\
  &\hspace{-2.08cm}=\sum_{i\in \mathcal{Y}}^{}\text{Pr}\left(Y_{\mathrm{t}}=j,Y_{\mathrm{c}}=i|X  \right)  \\
  &\hspace{-2.08cm}=\sum_{i\in \mathcal{Y}}^{}\text{Pr}\left(Y_{\mathrm{t}}=j|Y_{\mathrm{c}}=i,X  \right)\cdot \text{Pr}\left(Y_{\mathrm{c}}=i|X  \right).
\end{split}
\label{eq:p_Y_given_X}
\end{equation}
Based on the assumption from previous studies \cite{theory1, ntm1, ntm2, ntm3, ntm4, ntm5} that the transition from the clean labels $Y_{\mathrm{c}}$ to the noisy labels $Y_{\mathrm{t}}$ is independent of specific samples, then \cref{eq:p_Y_given_X} is reformulated as
\begin{equation}
\small
\text{Pr}\left(Y_{\mathrm{t}}=j|X \right)=\sum_{i\in \mathcal{Y}}^{}\text{Pr}\left(Y_{\mathrm{t}}=j|Y_{\mathrm{c}}=i  \right)\cdot \text{Pr}\left(Y_{\mathrm{c}}=i|X  \right).
\label{eq:p_Y_given_X_redefine}
\end{equation}
Substituting \cref{eq:p_Y_given_X_redefine} into 
 \cref{eq:risk_D_t_redefined} yields:
\begin{equation}
\small
\begin{split}
\mathcal{R}_{\tilde{D}_{\mathrm{t}}}\left( f \right) &  \\
  &\hspace{-1.2cm}= \mathbb{E}_{X\sim \tilde{D}_{\mathrm{t}}}\left[\sum_{j\in\mathcal{Y}}\left(\sum_{i\in\mathcal{Y}}T_{ij}\cdot\Pr(Y_{\mathrm{c}}=i\mid X)\right)\cdot\mathbb{I}(f(X)\neq j)\right] \\
  &\hspace{-1.2cm}=\mathbb{E}_{X\sim \tilde{D}_{\mathrm{t}}}\left[\sum_{i\in\mathcal{Y}}\Pr(Y_{\mathrm{c}}=i\mid X)\cdot\left(\sum_{j\in\mathcal{Y}}T_{ij}\cdot\mathbb{I}(f(X)\neq j)\right)\right].
\end{split}
\end{equation}
Based on the total probability law,
\begin{equation}
\small
\mathbb{E}_{X\sim \tilde{D}_{\mathrm{t}}}[\Pr(Y_{\mathrm{c}}=i\mid X)]=\Pr(Y_{\mathrm{c}}=i),
\end{equation}
the expectation can be shifted from $X$ to the distribution of $Y_{\mathrm{c}}$:
\begin{equation}
\small
\begin{split}
\mathcal{R}_{\tilde{D}_{\mathrm{t}}}(f)  &  \\
&\hspace{-1.1cm}=\sum_{i\in\mathcal{Y}}\Pr(Y_{\mathrm{c}}=i)\cdot\mathbb{E}_{X\sim \tilde{D}_{\mathrm{t}}}\Big[\sum_{i\in\mathcal{Y}}T_{ij}\cdot\mathbb{I}(f(X)\neq j)\mid Y_{\mathrm{c}}=i\Big].
\end{split}
\label{eq:r_D_t_redefine2}
\end{equation}
Let us dfine the confusion matrix $C(f)$, where $C_{ij}(f)=\text{Pr}(f(X)=j|Y_{\mathrm{c}}=i)$ represents the probability that classifier $f$ predicts label $j$ when the true label is $i$. The confusion matrix $C(f)$ satisfies for $\forall i,j\in \mathcal{Y},0\leqslant C_{ij}(f)\leqslant 1,\sum_{j}^{}C_{ij}(f)=1$. Therefore, the expectation of $\mathbb{I}(f(X)\neq j|Y_{\mathrm{c}}=i)$ can be rewritten as
\begin{equation}
\small
\mathbb{E}_{X\sim \tilde{D}_{\mathrm{t}}}[\mathbb{I}(f(X)\neq j)\mid Y_{\mathrm{c}}=i]=1-C_{ij}(f).
\label{eq:exp_x_D_t}
\end{equation}
Substituting \cref{eq:exp_x_D_t} into \cref{eq:r_D_t_redefine2} yields:
\begin{equation}
\small
\mathcal{R}_{\tilde{D}_{\mathrm{t}}}(f)=\sum_{i\in\mathcal{Y}}\Pr(Y_{\mathrm{c}}=i)\cdot\Big(\sum_{j\in\mathcal{Y}}T_{ij}\cdot(1-C_{ij}(f))\Big).
\end{equation}
Since $\sum_{j\in\mathcal{Y}}T_{ij}=1$, the accuracy of the classifier $f$ on the noisy distribution $\tilde{D}_{\mathrm{t}}$ can be solved via
\begin{equation}
\small
\begin{split}
\mathrm{Acc}_{\tilde{D}_{\mathrm{t}}}\left( f \right) & \\
  &\hspace{-1.5cm} =1-\mathcal{R}_{\tilde{D}_{\mathrm{t}}}(f)=\sum_{i\in\mathcal{Y}}\Pr(Y_{\mathrm{c}}=i)\cdot\left(\sum_{j\in\mathcal{Y}}T_{ij}\cdot C_{ij}(f)\right).
\end{split}
\end{equation}
Similarly, the accuracy for the clean distribution $\tilde{D}_{\mathrm{t}}$ is formulated as
\begin{equation}
\small
\mathrm{Acc}_{\tilde{D}_{\mathrm{c}}}\left( f \right)=1-\mathcal{R}_{\tilde{D}_{\mathrm{c}}}(f)=\sum_{i\in\mathcal{Y}}\Pr(Y_{\mathrm{c}}=i)\cdot C_{ii}(f).
\end{equation}
If the correct patterns can be learned from the noisy distribution, then the diagonal elements of the noise transition matrix $T$ has to be the largest in their corresponding rows, i.e., for $\forall i,j\in \mathcal{Y}, T_{ii} > T_{ij}, j\neq i$. As a result, 

\begin{equation}
\small
\sum_{i\in\mathcal{Y}}(\Pr(Y_{\mathrm{c}}=i)\sum_{j\in\mathcal{Y}}T_{ij}\cdot C_{ij}(f))\leqslant \sum_{i\in\mathcal{Y}}\Pr(Y_{\mathrm{c}}=i)\cdot T_{ii}.
\end{equation}
The equality holds if and only if $C(f)=I$, meaning that $f$ is the optimal classifier on the clean distribution. Therefore, the optimal classifiers on the noisy distribution and the corresponding clean distribution are consistent.

\section{}
Considering the sample size of the training set $D_{\mathrm{t}}$ is $n$, through random sampling, $D_{\mathrm{t}}$ can be divided into a sub-training set $\hat D_{\mathrm{t}}$ and a noisy validation set $\hat D_{\mathrm{v}}$ containing $n_{\mathrm{t}}$ and $n_{\mathrm{v}}$ samples, respectively. The sampling rate is denoted as $r=n_{\mathrm{t}}/n$. In NMC, we try to ensure that all the training data can be corrected through random sampling for multiple times. However, to achieve this goal, the sampling time would approach infinity. Therefore, this constraint is relaxed to $\beta$ of the samples in $D_{\mathrm{t}}$ being selected into $\hat D_{\mathrm{t}}$ at least once, where $\beta$ is a decimal close to 1. To solve the theoretical sampling times, the probability $P_{\mathrm{select}}$ that all samples in $D_{\mathrm{t}}$ are selected into $\hat D_{\mathrm{t}}$ at least once after sampling for $s$ times is first computed via
\begin{equation}
\small
P_{\text{select}}=\prod_{i=1}^{n}\left[ 1-\left( 1-r \right)^{s} \right].
\end{equation}
To ensure that at least $\beta$ of the training samples are selected into $\hat D_{\mathrm{t}}$, the following condition holds:
\begin{equation}
n\cdot \text{ln}\left[ 1-\left( 1-r \right)^{s} \right]\geqslant  \text{ln}\left( \beta \right).
\end{equation}
By exponentiating both sides, the logarithm is removed
\begin{equation}
1-\left( 1-r \right)^{s}\geqslant  \text{exp}\left[ \text{ln}\left( \beta \right)/n \right].
\end{equation}
Take the natural logarithm again and we can have:
\begin{equation}
s\cdot \text{ln}\left( 1-r \right)\leqslant \text{ln}\left[ 1-\text{exp}\left( \text{ln}\left( \beta \right)/n \right) \right].
\end{equation}
Therefore, when at least $\beta$ of the training samples are selected into $\hat D_{\mathrm{t}}$ at least once, the theoretical sampling time satisfies
\begin{equation}
s\geqslant \frac{\text{ln}\left( 1-\text{exp}\left( \text{ln}\left( \beta \right)/n \right) \right)}{\text{ln}\left( 1-r \right)}.
\end{equation}
The equation holds when the coverage rate of the selected samples over the training set equals $\beta$.

\end{document}